\documentclass[sn-mathphys,Numbered]{sn-jnl}% Math and Physical Sciences Reference Style
\usepackage{graphicx}%
\usepackage{multirow}%
\usepackage{amsmath,amssymb,amsfonts}%
\usepackage{amsthm}%
\usepackage{mathrsfs}%
\usepackage[title]{appendix}%
\usepackage{xcolor}%
\usepackage{textcomp}%
\usepackage{manyfoot}%
\usepackage{booktabs}%
\usepackage{algorithm}%
\usepackage{algorithmicx}%
\usepackage{algpseudocode}%
\usepackage{listings}%
\usepackage{adjustbox}
\usepackage{makecell}
\usepackage{threeparttable}
\usepackage{caption}
\usepackage{float}
\usepackage[numbers]{natbib}

\theoremstyle{thmstyleone}%
\theoremstyle{thmstyletwo}%

\theoremstyle{thmstylethree}%

\begin{document}

\title[Article Title]{PolypNextLSTM: A lightweight and fast polyp video segmentation network using ConvNext and ConvLSTM}


\author*[1]{\fnm{Debayan} \sur{Bhattacharya}}\email{debayan.bhattacharya@tuhh.de}
\equalcont{These authors contributed equally to this work.}

\author[1]{\fnm{Konrad} \sur{Reuter}}
\equalcont{These authors contributed equally to this work.}

\author[1]{\fnm{Finn} \sur{Behrendt}}

\author[1]{\fnm{Lennart} \sur{Maack}}

\author[1]{\fnm{Sarah} \sur{Grube}}

\author[1]{\fnm{Alexander} \sur{Schlaefer}}

\affil[1]{\orgdiv{Institute of Medical Technology and Intelligent Systems}, \orgname{Technische Universitaet Hamburg}, \city{Hamburg},  \country{Germany}}


%%================================%%
%% Sample for structured abstract %%
%%================================%%

 \abstract{\textbf{Purpose:} Commonly employed in polyp segmentation, single image UNet architectures lack the temporal insight clinicians gain from video data in diagnosing polyps. To mirror clinical practices more faithfully, our proposed solution, \textit{PolypNextLSTM}, leverages video-based deep learning, harnessing temporal information for superior segmentation performance with least parameter overhead, making it possibly suitable for edge devices.

 \textbf{Methods:} \textit{PolypNextLSTM} employs a UNet-like structure with ConvNext-Tiny as its backbone, strategically omitting the last two layers to reduce parameter overhead. Our temporal fusion module, a Convolutional Long Short Term Memory (ConvLSTM), effectively exploits temporal features. Our primary novelty lies in \textit{PolypNextLSTM}, which stands out as the leanest in parameters and the fastest model, surpassing the performance of five state-of-the-art image and video-based deep learning models. The evaluation of the SUN-SEG dataset spans easy-to-detect and hard-to-detect polyp scenarios, along with videos containing challenging artifacts like fast motion and occlusion.

 \textbf{Results:} Comparison against 5 image-based and 5 video-based models demonstrates \textit{PolypNextLSTM}'s superiority, achieving a Dice score of 0.7898 on the hard-to-detect polyp test set, surpassing image-based PraNet (0.7519) and video-based PNSPlusNet (0.7486). Notably, our model excels in videos featuring complex artefacts such as ghosting and occlusion.

 \textbf{Conclusion:} \textit{PolypNextLSTM}, integrating pruned ConvNext-Tiny with ConvLSTM for temporal fusion, not only exhibits superior segmentation performance but also maintains the highest frames per speed among evaluated models. Upon acceptance, the code for our model will be made accessible for broader utilization.}




\maketitle

\section{Introduction}\label{sec1}

Colorectal cancer stands as a significant concern, ranking as the second most common cancer among women and the third among men, contributing to approximately 10\% of global cancer cases. Its origin often traces back to the development of adenomatous polyps \cite{Pickhardt2018-ey}, emphasizing the criticality of early detection and removal to prevent cancer \citep{CRC_prevention, CRC_prevention_DE}.

%Colonoscopy remains the gold standard for detecting colorectal polyps, proving effective in reducing cancer incidence and mortality rates \citep{CRC_prevention, CRC_prevention_DE}. However, it's noteworthy that polyps below 5mm pose a challenge, with a miss rate exceeding 20\% \citep{Miss, Miss2}. Variations in size, shape, color, and surface structure further complicate detection, demanding expertise from gastroenterologists.

Deep learning-based polyp segmentation models may serve as secondary opinions for gastroenterologists, but limited labeled data from full-length colonoscopy videos poses a challenge \citep{Ahmad2021}. Clinical reports storing still frames create image-based polyp databases, enabling development of architectures like UNet, Vision Transformers, and Swin Transformers for segmentation \citep{Vázquez2017, 10.1007/978-3-030-37734-2_37, unet, zhou2018unet, jha2019resunet, YEUNG2021Focus, patel2021enhanced, dosovitskiy2021image, liu2021swin, zhang2021transfuse, dong2023polyppvt}. While current models mostly focus on single images, endoscopy units record video and thereby, image-based models do not leverage the temporal information to enhance segmentation. Image-based methods cannot contextualize within sequences, missing crucial context for accurate segmentation. Processing videos mirrors real-life scenarios and may ensure a more precise segmentation through the multi-view perspective of a suspected polyp \citep{Ahmad2021}. Another consideration to make is making models with less parameters so that they can still perform real-time inference. 

In the realm of video-based polyp segmentation, temporal information integration remains a relatively unexplored frontier. Puyal et al. proposed hybrid 2D/3D network \citep{hybrid2d3d}, where individual images undergo independent encoding by a shared backbone. Subsequently, 3D convolution layers amalgamate information across frames to yield segmentation outcomes. Building upon this, Ji et al. introduced the PNSNet architecture in 2021 \citep{ji2021progressively}, leveraging a novel 'Normalized Self-Attention Block' for temporal assimilation. Their subsequent iteration, PNS+, employs a global encoder which processes an anchor frame, a local encoder which processes subsequent frames and Normalized Self-Attention Block for improved performance \citep{Ji_2022}. Zhao et al. advanced the field by devising a semi-supervised network \citep{zhao2022semi}. This model employs multi-head attention modules separately for temporal and spatial dimensions, supplemented by an attention-based module during decoding. By reducing the need for laborious mask labeling, their approach aims to mitigate the time and effort involved in dataset annotation.

The limited research in video-based polyp segmentation is partly due to the scarcity of adequately large, densely labeled datasets. Ji et al. addressed this in 2022 by introducing the SUN-SEG dataset \citep{Ji_2022}, a restructured version of the SUN database \citep{SUN-data, SUN}. With meticulously created segmentation masks for positive cases across 1013 video clips,  158,690 images and defined training and test set splits, this dataset stands as the largest fully segmented resource available, serving as a benchmark for polyp video segmentation. Another reason for limited research is that computational efficiency poses a challenge. Models must strike a balance between lightweight design, high inference speed while having high segmentation performance. Image-based segmentation models, focusing on individual images, often display superior computational efficiency compared to their video-based counterparts. In response to these challenges, we propose the \textit{PolypNextLSTM}, a novel video polyp segmentation architecture. This framework integrates the latest ConvNext backbone \citep{ConvNext} and a bidirectional Convolutional Long Short Term Memory (ConvLSTM) module as our temporal fusion module. Notably, our model maintains the lowest parameter count among image and video-based state-of-the-art (SOTA) models while ensuring real-time processing capabilities. Our investigation delves into diverse temporal processing strategies beyond LSTM, considering computational cost, inference speed, and segmentation performance to inform our architectural choices. Overall, our main contributions are four-fold:

\begin{itemize}
    \item Introduction of the \textit{PolypNextLSTM} architecture, leveraging a pruned ConvNext-Tiny \citep{ConvNext} backbone integrated with a bidirectional convolutional LSTM to encapsulate temporal information making it the leanest model while still being the fastest and best performing model.
    
    \item We analyse the optimal video sequence length to process simultaneously.
    \item We explore the impact of different backbone architectures and temporal fusion modules and justify the reason for choosing ConvLSTM as a temporal fusion block.
    \item  We analyse the optimal placement of ConvLSTM, discerning its effects on overall performance metrics, inference speed, and model parameter count.

\end{itemize}

\section{Method}

\subsection{Dataset}

The SUN-SEG dataset, derived from the SUN-database \citep{SUN-data, SUN}, establishes a segmentation benchmark by meticulously crafting segmentation masks for each frame. Comprising originally of 113 videos, each video is segmented into smaller clips of 3-11 seconds each, at a frame rate of 30FPS—the dataset consists of 378 positive and 728 negative cases. Some of the smaller clips have polyps in the frame and some have no polyps, constituting 'positive' and 'negative' clips respectively. Only the positive polyp clips are used for the experiments. In the
training set there are often multiple clips which show the same polyp. The amount of
clips per polyp ranges from one to sixteen. To keep the amount of training data on a level that is easier to manage, only the first clip for each polyp is used. This leads to a training set of 51 clips of different polyps with a total of 9704 frames. The predefined test sets remain as they are.  The test set, categorized as SUN-SEG-Easy (119 clips, 17,070 frames) and SUN-SEG-Hard (54 clips, 12,522 frames), is entirely designated for testing, stratified by difficulty levels across pathological categories as outlined by the original work \citep{Ji_2022} as well as mentioned in their code repository \footnote{https://github.com/GewelsJI/VPS?tab=readme-ov-file}. Our predefined test sets, SUN-SEG-Easy and SUN-SEG-Hard, encompass two colonoscopy scenarios—'seen' and 'unseen'. 'Seen' delineates instances where the testing samples originate from the same case as the training set (33 clips in SUN-SEG-Easy, 17 clips in SUN-SEG-Hard). Conversely, 'unseen' indicates scenarios absent in the training set (86 clips in SUN-SEG-Easy, 37 clips in SUN-SEG-Hard), enabling a more comprehensive evaluation of model performance under distinct conditions.

The SUN-SEG database offers another advantage. All clips are labeled with visual attributes that occur in it. Splitting results by visual attributes allows for a more in-depth analysis and can help to identify strength and weaknesses of models. All possible visual attributes and a description are listed in table \ref{tab:visual_attributes}.
\captionsetup[table]{position=top}

\begin{table}[t]
    \centering
    \captionsetup{justification=justified, singlelinecheck=false, position=top}

    \begin{adjustbox}{width=\textwidth}
    \begin{tabular}{|c|c|c|}
        \hline
        \textbf{ID} & \textbf{Name} & \textbf{Description} \\
        \hline
        SI & Surgical Instruments & \makecell{The endoscopic surgical procedures involve the positioning of\\ instruments, such as snares, forceps, knives and electrodes.}\\
        \hline
        IB & Indefinable Boundaries & \makecell{The foreground and background areas around the object have a \\similar colour.} \\
        \hline
        HO & Heterogeneous Object & Object regions have distinct colours. \\
        \hline
        GH & Ghosting & \makecell{Object has anomaly RGB-colored boundary due to fast-moving \\or insufficient refresh rate.} \\
        \hline
        FM & Fast Motion & \makecell{The average per-frame object motion, computed as the Euclidean \\distance of polyp centroids between consecutive frames,\\ is larger than 20 pixels.}\\
        \hline
        SO & Small Object & \makecell{The average ratio between the object size and the image area\\ is smaller than 0.05.} \\
        \hline
        LO & Large Object & \makecell{The average ratio between the object bounding-box area and\\ the image area is larger than 0.15.}\\
        \hline
        OCC & Occlusion & Object becomes partially or fully occluded. \\
        \hline
        OV & Out of View & Object is partially clipped by the image boundaries. \\
        \hline
        SV & Scale Variation & \makecell{The average area ratio among any pair of bounding boxes\\ enclosing the target object is smaller than 0.5.} \\
        \hline
    \end{tabular}
    \end{adjustbox}
    \caption{Overview on the visual attribute labels in the SUN-SEG database. Descriptions are copied from the official Git repository \href{https://github.com/GewelsJI/VPS}{https://github.com/GewelsJI/VPS}}. 
    
    \label{tab:visual_attributes}
\end{table}

\subsection{Proposed method}

\begin{figure}[H]
    \centering
    \includegraphics[width=\textwidth]{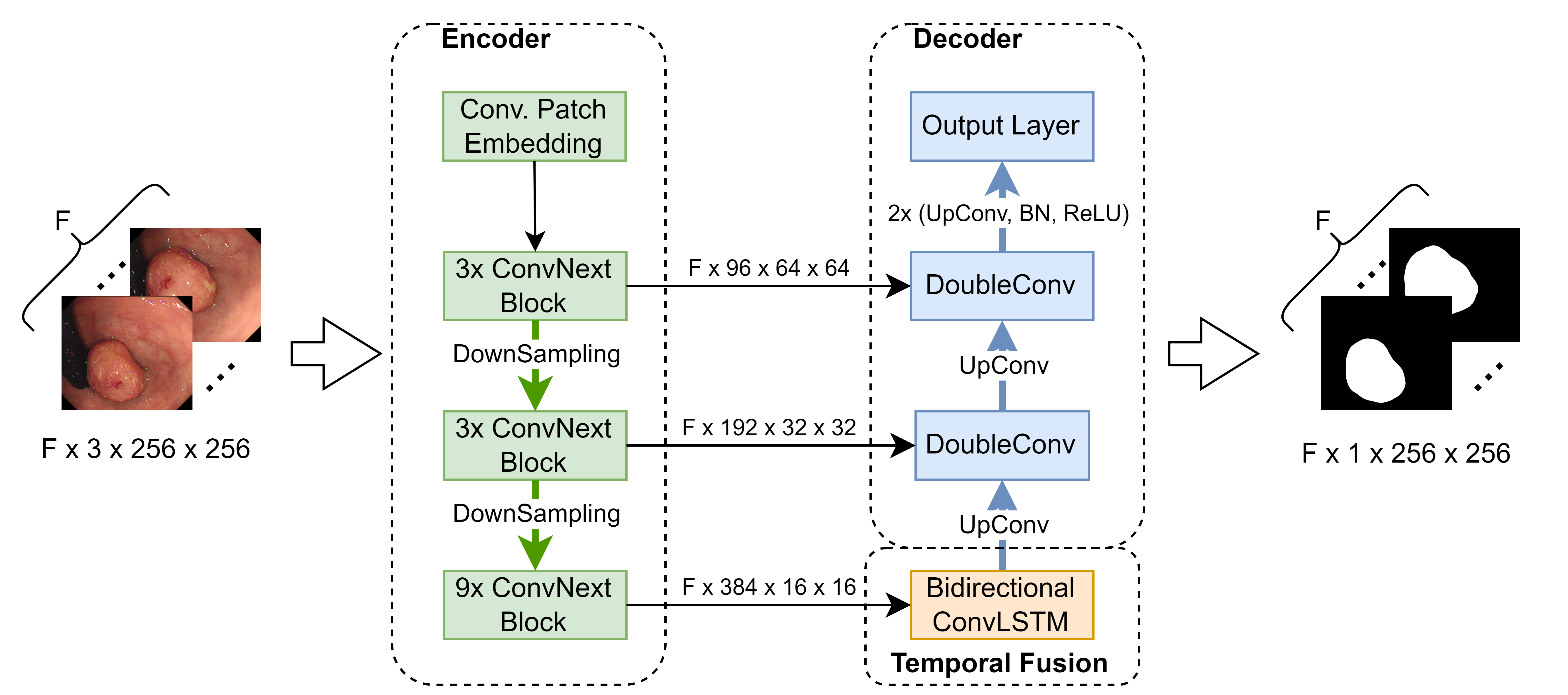}
    \caption[Proposed Model]{The proposed Model. A reduced ConvNext-tiny is used as the encoder. The information between the encoded frames is fused using a bidirecitonal ConvLSTM. The decoder is inspired by the UNet. \(F\) is the number of subsequent frames being processed simultaneously by the model. }
    \label{fig:our_model}
\end{figure}

\begin{figure}[h]
    \centering
    \includegraphics[width=\textwidth]{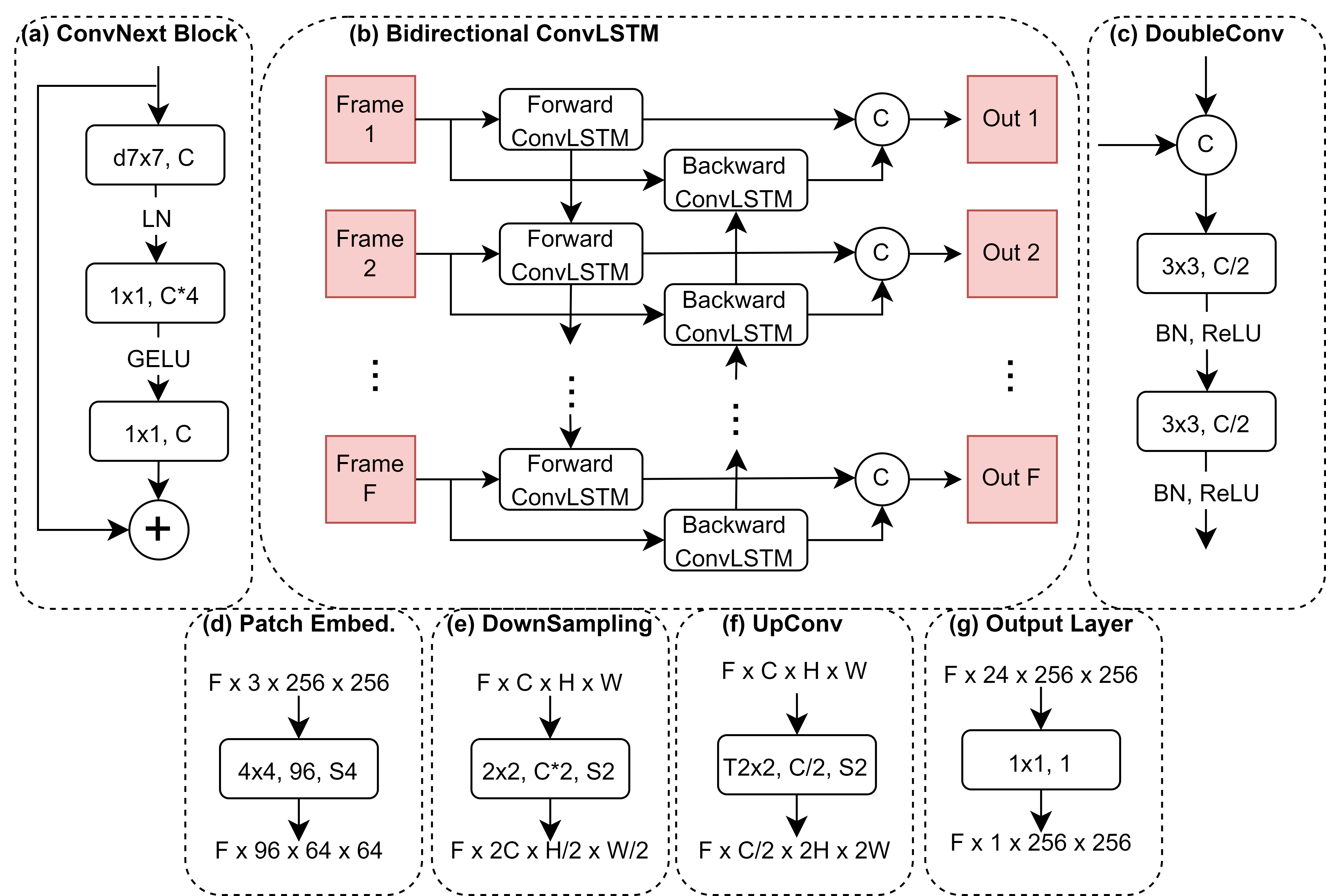}
    \caption[Proposed Model: Components]{Key components of our network: 
(a) ConvNextBlock - main encoder building block.  (b) Bidirectional ConvLSTM - fuses information across frames.  (c) DoubleConv module - merges skip connection data with upsampled information.  (d) Patch embedding layer - serves as the encoder's input. 
(e) Downsampling layers and (f) Upsampling layers.  (g) Output layer - reduces channel dimension to 1.}
    \label{fig:components}
\end{figure}

We propose a new video polyp segmentation network which is based on a ConvNext-tiny backbone and uses a bidirectional convolutional LSTM to incorporate temporal information. The proposed model is shown in figure \ref{fig:our_model}. The different components are explained in more detail in figure \ref{fig:components}. 

 We opt for the ConvNext-tiny as our backbone due to its recent advancements in convolutional neural networks, striking a balance between precision and efficiency. This model refines ResNet by integrating design elements like grouped convolution, inverted bottleneck, larger kernel sizes, and micro designs at the layer level \citep{ConvNext}. Our customized version, termed 'reduced ConvNext-tiny,' is achieved by eliminating the classification layers and the final downsampling stage, resulting in a more lightweight model. Unlike the original four-stage ConvNext-tiny with (3, 3, 9, 3) ConvNext blocks, our reduced backbone operates with three stages containing (3, 3, 9) ConvNext blocks. By omitting the parameter-heavy final blocks, we significantly trim down the model parameters, reducing from 27.82 million to 12.35 million.

The ConvNext block (depicted in figure \ref{fig:components} (a)) comprises a depthwise $7 \times 7$ convolution followed by two $1 \times 1$ convolutions. Additionally, our architecture employs a bidirectional convolutional LSTM (illustrated in figure \ref{fig:components} (b)) to fuse information across consecutive frames, operating in a 'many-to-many' i.e. \(F\) segmentation mask generated for \(F\) images. This ConvLSTM maintains the input channel integrity despite halving the channel count from $384$ to $192$ during convolution by concatenating the forward and backward features.

Drawing inspiration from the UNet decoder, our model gradually upsamples images while incorporating earlier encoder data through skip connections. The DoubleConv block (shown in figure \ref{fig:components} (c)) consists of two $3 \times 3$ convolutions with batch normalization and ReLU activation, halving the channel dimensions. Upsampling is achieved using a $2 \times 2$ transposed convolution with a stride of $2$ (exemplified in figure \ref{fig:components} (f)), ultimately reduced to a $1 \times 1$ convolutional output layer (seen in figure \ref{fig:components} (g)) for final segmentation masks, reducing the channel dimension from 24 to 1.

Our network expects input of shape $B \times F \times C \times H \times W$—batch dimension ($B$), frame sequence length ($F$), channel count ($C$), image height ($H$), and width ($W$). Data processing within the encoder and decoder involves flattening the batch and frame dimensions into one dimension, ensuring independent image processing. Only in the temporal fusion module is the data processed in its original form.

\subsection{Implementation Detail} 

We conducted all model training on a system equipped with an AMD Ryzen 9 3950X CPU, an Nvidia RTX 3090 graphics card, and 64 GBs of RAM, employing PyTorch 1.13 as the deep learning framework. Adhering to PyTorch's reproducibility guidelines \footnote{https://pytorch.org/docs/1.13/notes/randomness.html}, we ensure the replicability of all experiments without variance in results. Our chosen configuration includes a temporal dimension of 5 consecutive frames (\(F\)) and a batch size (\(B\)) of 8. In the image-based model segments—both encoder and decoder—the batch and temporal dimensions are flattened into one, effectively creating a batch size of 40. Consequently, inputs for modules involving the temporal dimension take the shape of $8 \times 5 \times C \times H \times W$, where $C$ signifies channel count and $H$ \& $W$ denote height and width, while the encoder and decoder operate on data of shape $40 \times C \times H \times W$, all images standardized to a fixed size of $256 \times 256$. Augmentation techniques include random rotations, horizontal and vertical flips, and random center cropping, consistently applied to the five input images.

For fairness in comparisons, we use a batch size of 8 and 5 consecutive frames for all SOTA. To align COSNet with other models, we adjust its input by using the first and last frames from sets of 5 frames. When temporal information is not used, we flatten batch and frame dimensions for an effective size of 40. Deep supervision techniques are applied per the authors' recommendations.

We utilized Adam as the optimizer with an initial learning rate of 1e-4. The loss function is a fusion of Dice loss and binary cross-entropy loss. Our experiments entail 5-fold cross-validation across 100 epochs. All models are evaluated using four common segmentation metrics: Dice score, intersection
over union (IOU), 95\% Hausdorff distance (HD95) and recall.

\section{Results}

\subsection{Comparison to state-of-the-art image and video models}

Tables  \ref{tab:sota_unseen} display the comprehensive performance evaluations of various methods on SUN-SEG-Hard test sets, categorized as 'Easy Unseen' and 'Hard Unseen'. Our model consistently surpasses all comparative models across all metrics, including 'Seen' and 'Unseen' scenarios.

In Table \ref{tab:sota_unseen}, PraNet emerges as the second-best performer across most metrics, excluding HD95, where HybridNet secures the second spot. Notably, \textit{PolypNextLSTM} shows better performance on the 'Hard Unseen' test set compared to the 'Easy Unseen'. There is an improvement on the 'Easy Unseen' test set with +0.0129 (+1.71\%) dice score, +0.0131 (+1.92\%) IOU, -1.34 (-7.77\%) Hausdorff distance, and +0.0152 (+2.1\%) recall in comparison to PraNet for each metric. The improvement on the 'Hard Unseen' test set is even more substantial, with +0.0319 (+4.24\%) dice score, +0.0307 (+4.54\%) IOU, -1.59 (-10.2\%) Hausdorff distance, and +0.0323 (+4.41\%) recall, indicating our approach's proficiency in detecting challenging polyps.

Furthermore, our model outperforms both image and video state-of-the-art models while utilizing the fewest parameters and exhibiting the highest inference speed. The Frames Per Second (FPS) metric, evaluating the processing speed for a video snippet of five frames at a resolution of $256\times256$ pixels, illustrates our model's efficiency.

\begin{threeparttable}[H]
    \centering

    \begin{adjustbox}{width=\textwidth}
    \begin{tabular}{|c|l|c|c|c|c|c|c|c|c|c|c|}
        \hline
        {} & {} &  \multicolumn{4}{c|}{Easy Unseen} & \multicolumn{4}{c|}{Hard Unseen} & \multicolumn{2}{c|}{}\\
        {} & {} & Dice & IOU & HD95 & Recall & Dice & IOU & HD95 & Recall & Params & FPS\\
        \hline
        \multirow{5}{*}{\rotatebox[origin=c]{90}{\textbf{Image}}} & DeepLab \citep{deeplabV3} & 0.7046 & 0.6196 & 22.28 & 0.6483 & 0.7107 & 0.6214 & 19.57 & 0.6651 & 39.63M & 54 \\
        & PraNet \citep{pranet} & 0.7557 & 0.6827 & 17.52 & 0.7198 & 0.7519 & 0.6760 & 15.96 & 0.7318 & 32.55M & 45\\
        & SANet \citep{SANet} & 0.7412 & 0.6638 & 18.34 & 0.6951 & 0.7465 & 0.6624 & 17.12 & 0.7157 & 23.90M & 71 \\
        & TransFuse \citep{zhang2021transfuse} & 0.7058 & 0.6225 & 23.26 & 0.6549 & 0.6804 & 0.5973 & 24.84 & 0.6414 & 26.27M & 63 \\
        & CASCADE \citep{CASCADE} & 0.7419 & 0.6672 & 19.23 & 0.7042 & 0.7170 & 0.6393 & 20.28 & 0.6938 & 35.27M & 54 \\
        \hline
        \multirow{5}{*}{\rotatebox[origin=c]{90}{\textbf{Video}}} & COSNet \citep{cosnet} & 0.6574 & 0.5761 & 27.01 & 0.6083 & 0.6427 & 0.5598 & 26.02 & 0.6085 & 81.23M & 16 \\
        & HybridNet \citep{hybrid2d3d} & 0.7350 & 0.6492 & 17.25 & 0.7013 & 0.7214 & 0.6334 & 15.66 & 0.7070 & 101.5M & 67\\ 
        & PNSNet \citep{ji2021progressively} & 0.7313 & 0.6474 & 21.00 & 0.6805 & 0.7392 & 0.6526 & 17.97 & 0.7052 & 26.87M & 61 \\
        & PNSPlusNet \citep{Ji_2022} & 0.7422 & 0.6647 & 19.00 & 0.7010 & 0.7486 & 0.6660 & 16.11 & 0.7266 & 26.87M & 57 \\
        & SSTAN \citep{zhao2022semi} & 0.7157 & 0.6363 & 23.40 & 0.6760 & 0.6964 & 0.6163 & 24.05 & 0.6740 & 30.15M & 101 \\
        \hline
        & \textbf{Ours} & \textbf{0.7686}  & \textbf{0.6958}  & \textbf{15.91}  & \textbf{
        0.7350}  & \textbf{0.7838}  & \textbf{0.7067}  & \textbf{14.07}  & \textbf{0.7641} & \textbf{21.95M} & \textbf{108}\\
        \hline
    \end{tabular}
    \end{adjustbox}
    \caption[Comparison to State-of-the-Art Models: Unseen Cases]{Comparison to various state-of-the-art models on the unseen cases. The top five models are image models, while the bottom five are video-based models.}
    \label{tab:sota_unseen}
\end{threeparttable}

We also present the Dice score results categorized by visual attributes (see Table \ref{tab:visual_attributes}) in Table \ref{tab:attributes_eu} for the 'Easy Unseen' test set and in Table \ref{tab:attributes_hu} for the 'Hard Unseen' test set. In the 'Easy Unseen' set, our model excels in multiple attributes—HO, GH, FM, OV, and SV. Notably, our model demonstrates significant improvement in SV, achieving +0.0255 (+4.04\%) compared to the second-best model, PraNet, in this category. While our model performs competitively in other categories, the largest margin appears in LO, where PraNet outperforms by +0.0266 (+3.60\%). In the 'Hard Unseen' test set, our model emerges as the top performer across all categories. Particularly noteworthy is the substantial improvement in IB, showcasing +0.0321 (+5.24\%) compared to the second-best model (CASCADE). Given the generally lower scores, IB stands out as the most challenging category. 

\begin{threeparttable}[H]
    \centering
    \begin{adjustbox}{width=\textwidth}
    \begin{tabular}{|c|l|c|c|c|c|c|c|c|c|c|c|}
        \hline
        {} & {} & SI & IB & HO & GH & FM & SO & LO & OCC & OV & SV \\
        \hline
        \multirow{5}{*}{\rotatebox[origin=c]{90}{\textbf{Image}}} &DeepLab\citep{deeplabV3} & 0.7081 & 0.4844 & 0.8227 & 0.7382 & 0.6130 & 0.5400 & 0.6743 & 0.6391 & 0.6722 & 0.5840\\
        &PraNet\citep{pranet} & \textbf{0.7746} & 0.5490 & 0.8659 & 0.7867 & 0.6501 & 0.5979 & \textbf{0.7651} & \textbf{0.7155} & 0.7244 & 0.6317\\
        &SANet\citep{SANet} & 0.7683 & 0.5332 & 0.8471 & 0.7827 & 0.6392 & 0.5684 & 0.7444 & 0.6977 & 0.7126 & 0.6183\\
        &TransFuse\citep{zhang2021transfuse} & 0.6566 & 0.5292 & 0.7859 & 0.7262 & 0.6367 & 0.6005 & 0.6214 & 0.6201 & 0.6584 & 0.5594\\
        &CASCADE\citep{CASCADE} & 0.7111 & \textbf{0.5888} & 0.8510 & 0.7559 & 0.6455 & \textbf{0.6243} & 0.6753 & 0.6724 & 0.6826 & 0.6067\\
        \hline
        \multirow{5}{*}{\rotatebox[origin=c]{90}{\textbf{Video}}} &COSNet\citep{cosnet} & 0.6306 & 0.4277 & 0.7684 & 0.7073 & 0.5887 & 0.4880 & 0.6062 & 0.6106 & 0.6051 & 0.5093 \\
        &HybridNet\citep{hybrid2d3d} & 0.7554 & 0.4973 & 0.8687 & 0.7875 & 0.6376 & 0.5447 & 0.7505 & 0.7109 & 0.7307 & 0.6006\\
        &PNSNet\citep{ji2021progressively} & 0.7415 & 0.5417 & 0.8504 & 0.7511 & 0.6163 & 0.6108 & 0.7073 & 0.6852 & 0.6916 & 0.6114\\
        &PNSPlusNet\citep{Ji_2022} & 0.7467 & 0.5272 & 0.8700 & 0.7742 & 0.6319 & 0.5974 & 0.7244 & 0.6874 & 0.7144 & 0.6300\\
        &SSTAN\citep{zhao2022semi} & 0.7095 & 0.4946 & 0.8428 & 0.7598 & 0.6248 & 0.5562 & 0.6837 & 0.6616 & 0.6797 & 0.5931\\
        \hline
        & \textbf{Ours} & 0.7510 & 0.5704 & \textbf{0.8837} & \textbf{0.7973} & \textbf{0.6638} & 0.6225 & 0.7385 & 0.7101 & \textbf{0.7337} & \textbf{0.6572}\\
        \hline
    \end{tabular}
    \end{adjustbox}
    \caption[Dice by Visual Attribute: Easy Unseen]{Comparison of the dice score divided by the visual attributes occurring in the clips of the "Easy Unseen" test set. The best score for each category is marked in bold.}
    \label{tab:attributes_eu}
\end{threeparttable}

\begin{threeparttable}[H]
    \centering
    \begin{adjustbox}{width=\textwidth}
    \begin{tabular}{|c|l|c|c|c|c|c|c|c|c|c|c|}
        \hline
        {} & {} & SI & IB & HO & GH & FM & SO & LO & OCC & OV & SV \\
        \hline
        \multirow{5}{*}{\rotatebox[origin=c]{90}{\textbf{Image}}} &DeepLab\citep{deeplabV3} & 0.6984 & 0.5524 & 0.7519 & 0.7157 & 0.7131 & 0.6686 & 0.7583 & 0.7261 & 0.7453 & 0.6473\\
        &PraNet\citep{pranet} & 0.7709 & 0.5865 & 0.8247 & 0.7588 & 0.7342 & 0.7027 & 0.8250 & 0.7796 & 0.7847 & 0.6955\\
        &SANet\citep{SANet} & 0.7518 & 0.6011 & 0.8048 & 0.7464 & 0.7394 & 0.7011 & 0.8042 & 0.7755 & 0.7759 & 0.6831\\
        &TransFuse\citep{zhang2021transfuse} & 0.6287 & 0.5722 & 0.6537 & 0.6602 & 0.7274 & 0.6675 & 0.5848 & 0.6557 & 0.6697 & 0.6149\\
        &CASCADE\citep{CASCADE} & 0.6769 & 0.6125 & 0.7231 & 0.6969 & 0.7507 & 0.6889 & 0.6899 & 0.7194 & 0.7307 & 0.6524\\
        \hline
        \multirow{5}{*}{\rotatebox[origin=c]{90}{\textbf{Video}}} &COSNet\citep{cosnet} & 0.6103 & 0.4801 & 0.6720 & 0.6303 & 0.6701 & 0.6037 & 0.6107 & 0.6463 & 0.6473 & 0.5880\\
        &HybridNet\citep{hybrid2d3d} & 0.7257 & 0.5307 & 0.8102 & 0.7248 & 0.7131 & 0.6312 & 0.8130 & 0.7653 & 0.7712 & 0.6557\\
        &PNSNet\citep{ji2021progressively} & 0.7482 & 0.5901 & 0.7879 & 0.7400 & 0.7298 & 0.7162 & 0.7663 & 0.7569 & 0.7615 & 0.6970\\
        &PNSPlusNet\citep{Ji_2022} & 0.7567 & 0.6026 & 0.8047 & 0.7565 & 0.7381 & 0.7165 & 0.7812 & 0.7693 & 0.7778 & 0.7070\\
        &SSTAN\citep{zhao2022semi} & 0.6721 & 0.5207 & 0.7405 & 0.6878 & 0.7184 & 0.6444 & 0.7232 & 0.7131 & 0.7230 & 0.6376\\
        \hline
        & \textbf{Ours} & \textbf{0.8000} & \textbf{0.6446} & \textbf{0.8461} & \textbf{0.7678} & \textbf{0.7693} & \textbf{0.7318} & \textbf{0.8326} & \textbf{0.7984} & \textbf{0.8153} & \textbf{0.7139}\\
        \hline
    \end{tabular}
    \end{adjustbox}
    \caption[Dice by Visual Attribute: Hard Unseen]{Comparison of the dice score divided by the visual attributes occurring in the clips of the "Hard Unseen" test set. The best score for each category is marked in bold.}
    \label{tab:attributes_hu}
\end{threeparttable}

\begin{figure}[H]
    \centering
    \includegraphics[height=.75\textheight]{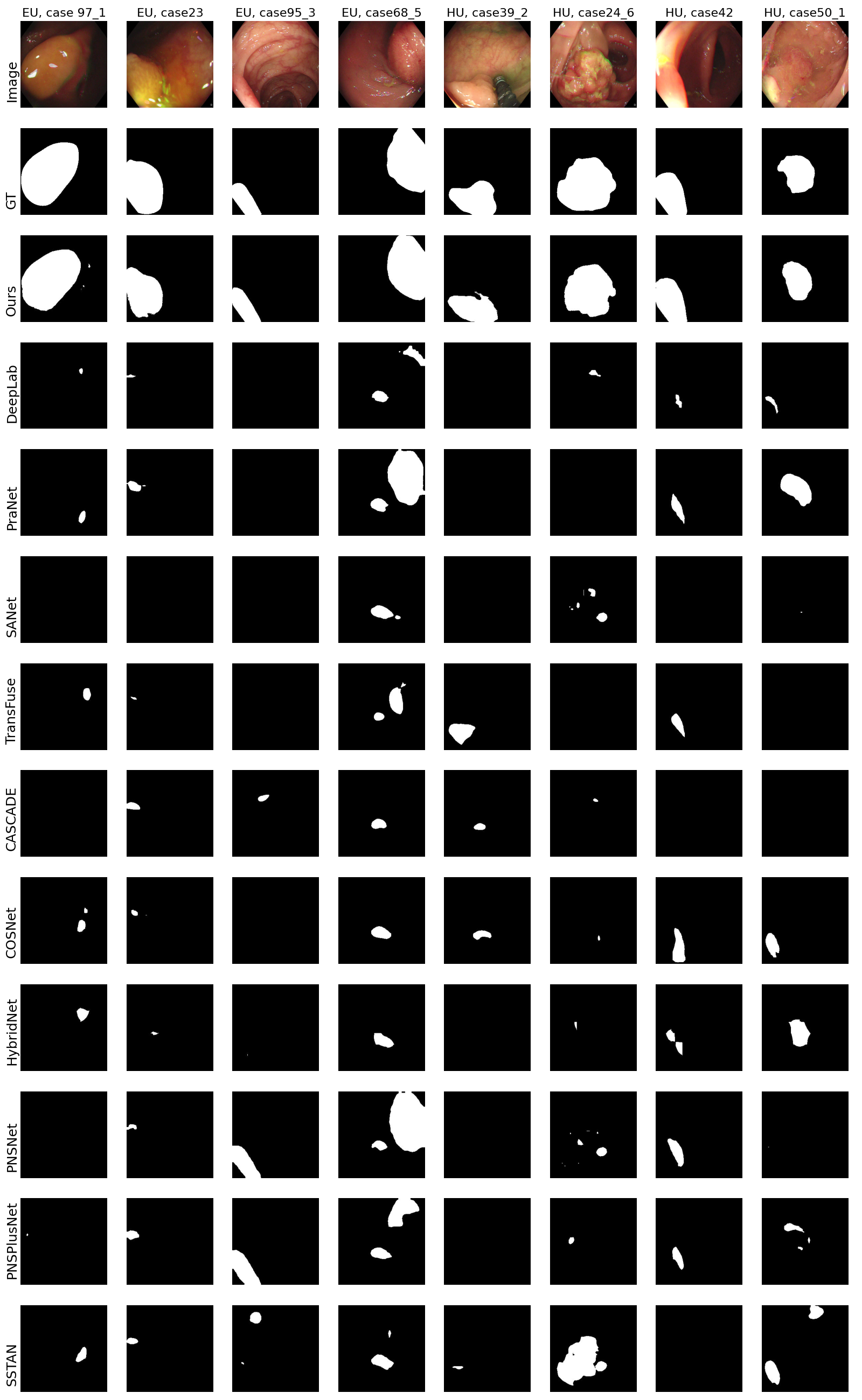}
    \caption[Qualitative Results]{Example results for cases where our model performed considerably better than other state-of-the-art models. The left four images are from the "Easy Unseen" test set and right frames from the "Hard Unseen" test set.}
    \label{fig:qualitative}
\end{figure}

In figure \ref{fig:qualitative}, we qualitatively compare the SOTA and our proposed model. On the left side there are the results for four cases from the "Easy Unseen" (EU) test set and on the right side for examples from the "Hard Unseen" (HU) test set. The case numbers are taken from the SUN-SEG dataset. 

Experiments with 'Seen' test set configuration are presented in Online Resource section 1.1.  

\subsection{Number of frame variation ablation study} 

\begin{figure}[H]
    \centering
    \includegraphics[width=\textwidth]{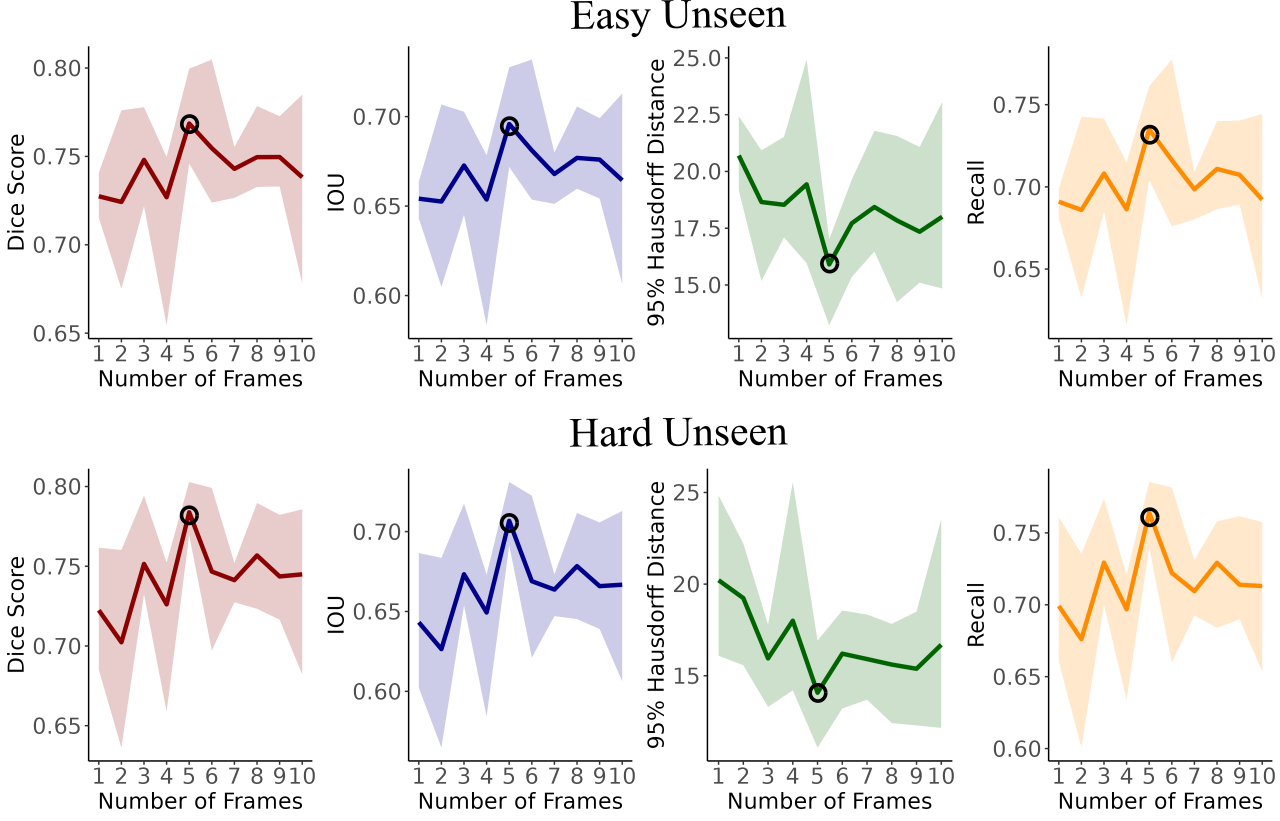}
    \caption[Number of Frames Variation:  Easy Unseen and Hard Unseen]{Variation of the number of frames for the different test set configurations for four different metrics. The coloured interval refers to the minimum and maximum of the cross-validation. Black circle shows the highest metric.}
    \label{fig:n_frames_eu}
\end{figure}

We investigate the impact of varying input frames on our proposed \textit{PolypNextLSTM}. Figure \ref{fig:n_frames_eu} showcases the results across different metrics concerning the number of frames for all test set configurations. \textit{PolypNextLSTM} exhibit their poorest performance with one or two input frames, gradually improving up to five frames where a distinct performance peak emerges across all metrics. Beyond five frames, there is a noticeable decline in results. Hence, empirically, processing five frames emerges as the optimal configuration where \textit{PolypNextLSTM} delivers its peak performance across both 'Seen' and 'Unseen' test set configurations. For results with the 'Seen' test set as well as tabular format of the aforementioned results, please refer to the Online Resource Section 1.2.

\section{Discussion} 

Comparisons with SOTA image and video-based segmentation networks (Tables \ref{tab:sota_unseen}) consistently position \textit{PolypNextLSTM} as the frontrunner. Its performance excels particularly in 'Unseen' scenarios, indicating strength in resembling clinical settings with new patients. We extensively justify the choice of ConvNext as a backbone in our Online Resource Section 2.1. Our investigation explores optimal ConvLSTM placements within skip connections (Online Resource Section 2.4) and the encoder network (Online Resource Section 2.5). Furthermore, we delve into various temporal fusion modules, including channel stacking, 3D convolutions akin to HybridNet \citep{hybrid2d3d}, unidirectional ConvLSTM, multi-headed attention, and normalized self-attention from PNSNet \citep{ji2021progressively} and PNSPlusNet \citep{Ji_2022}. Detailed analysis in Online Resource Section 2.3 concludes that the bidirectional ConvLSTM at the bottleneck ensures optimal performance without compromising computational efficiency or throughput. Models built around ConvLSTM exhibit adaptability to varying input sequence lengths without escalating parameters, distinguishing them from convolution-based approaches (channel stacking, 3D convolution) that inflate parameters with sequence length, affecting speed and ease of training.

 While PraNet stands out among image-based models, our temporal information integration outperforms it with nearly 50\% fewer parameters and over double the FPS. Surprisingly, models perform better on the 'hard' test set, possibly due to a training set bias towards tougher cases. An attribute-based analysis on the 'Easy Unseen' test set (table \ref{tab:attributes_eu}) indicates our method's strength across various attributes, especially in handling heterogeneous objects, ghosting, fast motion, out-of-view instances, and scale variation. Notably, scale variation witnesses a significant +0.0255 (+4.04\%) Dice score improvement compared to the next-best approach (PraNet). Temporal information proves beneficial for ghosting and out-of-view cases, leveraging multiple frames for better predictions despite visual artifacts. While our model consistently performs above average, challenges surface in segmenting large objects, where PraNet outperforms, possibly due to the network's restricted depth arising from certain ConvNext-tiny backbone layer removals. Intriguingly, results on the 'Hard Unseen' test set categorized by visual attributes (table \ref{tab:attributes_hu}) reveal our model's dominance across all categories, reinforcing the bidirectional ConvLSTM's role in precise segmentation through effective multi-frame information fusion.

Varying the number of frames indicates suboptimal results for sequences of one or two frames (figure \ref{fig:n_frames_eu}). This implies crucial information spanned across multiple frames, effectively incorporated by the bidirectional ConvLSTM into the segmentation process. However, longer sequences do not consistently yield better results, with a peak observed at five frames and subsequent stagnation or decline. While ConvLSTMs theoretically handle longer sequences, our results suggest that varying the sequence length could be a crucial hyperparameter linked to dataset complexity. For instance, PNSPlusNet \citep{Ji_2022} utilizes one anchor frame and five randomly selected subsequent frames for the same dataset which is similar to our five frames peak performance.

Although our study exhibits strong performance, it has limitations. Primarily, we have tested our method solely on one video polyp segmentation dataset. To establish its robustness and generalizability, future work should evaluate this model across multiple image and video polyp datasets. Additionally, as this study is retrospective, a prospective study would provide more accurate insights into its true performance. Despite these limitations, our \textit{PolypNextLSTM} stands out as the most lightweight and high-performing video-based polyp segmentation model. Its open-source implementations pave the way for further advancements in this domain.

\section{Conclusion}
We devised \textit{PolypNextLSTM}, an architecture employing ConvNext-Tiny \citep{ConvNext} as the backbone, integrated with ConvLSTM for temporal fusion within the bottleneck layer. Our model not only delivers superior segmentation performance but also maintains the highest FPS among the evaluated models. Evaluations conducted on the SUN-SEG dataset, the largest video polyp segmentation dataset to date, provide comprehensive insights across various test set scenarios. %Additionally, our approach aligns with the concept of independently encoding images and establishing connections on fully encoded states, consistent in four out of five compared video-based models \citep{cosnet, hybrid2d3d, ji2021progressively, Ji_2022}.

\section*{Conflict of Interest}
The authors state no conflict of interest. 

\section*{Ethical statement}
The research conducted for this paper adheres to ethical principles and guidelines concerning the utilization of publicly available datasets. The dataset employed in this study, SUN-SEG is a publicly accessible resource without individual identifiers, thus obviating the need for specific consent from individuals.

\bibliography{sn-bibliography}% common bib file


\end{document}